\documentclass[10pt,twocolumn,letterpaper]{article}
\usepackage[pagenumbers]{kt} 

\usepackage{graphicx}
\usepackage{amsmath}
\usepackage{amssymb}
\usepackage{booktabs}
\usepackage{array}

\usepackage[ruled,linesnumbered]{algorithm2e}
\usepackage[T1]{fontenc}
\usepackage{multirow}
\usepackage{float}

\usepackage[pagebackref,breaklinks,colorlinks]{hyperref}

\usepackage[capitalize]{cleveref}
\crefname{section}{Sec.}{Secs.}
\Crefname{section}{Section}{Sections}
\Crefname{table}{Table}{Tables}
\crefname{table}{Tab.}{Tabs.}

\newcommand{\cznet}{KT-Net}

\newcommand{\assistanta}{recovery assistant}
\newcommand{\Assistanta}{Recovery Assistant}

\newcommand{\Assistantb}{Discrimination Assistant}

\begin{document}

\title{KT-Net: Knowledge Transfer for Unpaired 3D Shape Completion}

\author{Zhen Cao\textsuperscript{1*}, Wenxiao Zhang\textsuperscript{1*}, 
Xin Wen\textsuperscript{2}\footnotemark[2], Zhen Dong\textsuperscript{1}\footnotemark[2], 
Yu-shen Liu\textsuperscript{3}, Xiongwu Xiao\textsuperscript{1}, Bisheng Yang\textsuperscript{1} \\
\textsuperscript{1}Wuhan University, \textsuperscript{2}JD.com,
\textsuperscript{3}Tsinghua University \\
{\tt\small \{zhen.cao,dongzhenwhu,xwxiao,bshyang\}@whu.edu.cn, wenxxiao.zhang@gmail.com} \\
{\tt\small wenxin16@jd.com, liuyushen@tsinghua.edu.cn}
}
\maketitle

\begin{abstract}
Unpaired 3D object completion aims to predict a complete 3D shape from an incomplete input without knowing the correspondence between the complete and incomplete shapes. In this paper, we propose the novel \cznet~ to solve this task from the new perspective of knowledge transfer. \cznet~ elaborates a teacher-assistant-student network to establish multiple knowledge transfer processes. Specifically, the teacher network takes complete shape as input and learns the knowledge of complete shape. The student network takes the incomplete one as input and restores the corresponding complete shape. And the assistant modules not only help to transfer the knowledge of complete shape from the teacher to the student, but also judge the learning effect of the student network. As a result, \cznet~ makes use of a more comprehensive understanding to establish the geometric correspondence between complete and incomplete shapes in a perspective of knowledge transfer, which enables more detailed geometric inference for generating high-quality complete shapes. We conduct comprehensive experiments on several datasets, and the results show that our method outperforms previous methods of unpaired point cloud completion by a large margin.
\end{abstract}

\section{Introduction}
Point clouds, as a popular representation form of 3D geometries, have drawn a growing research concern in many fields, such as computer vision \cite{qi2017pointnet}, robotics \cite{liu2015robotic}
and auto-navigation \cite{yue2018lidar}. However, the raw point clouds obtained by 3D scanning devices are often sparse, noisy and incomplete, which requires pre-processing (e.g. completion, denoising) before the downstream tasks. In this paper, we focus on the specific task to predict the complete shape for an incomplete point cloud, where the missing shape is usually caused by the limited scanning view and occlusion. Previous deep learning based studies in this field \cite{yuan2018pcn,liu2020morphing,huang2020pf,xie2020grnet} usually rely on the paired supervision of incomplete shape and its corresponding complete ground truth. However, in spite of the great progress achieved by these supervised methods, the completion performance is still limited by the insufficiency of paired training data, which is caused by the difficulty of obtaining the complete ground truth for incomplete shape in real-world scenarios.

A promising solution to this problem is to train the completion network in an unpaired manner. That is, the network is only fed by collections of complete and incomplete shapes without paired correspondence. Such intuition is driven by the easy access to the incomplete scanning in the real-world, and the large amount of complete shapes from many large-scaled 3D shape datasets \cite{chang2015shapenet, pan2021variational}. 
However, a typical challenge of unpaired shape completion task is the absence of the one-to-one strong supervision for each incomplete shape. 
Previous methods (e.g. Cycle4Completion \cite{wen2021cycle4completion} and Pcl2Pcl \cite{chen2019unpaired}) usually consider the adversarial global feature matching framework as the basic solution to this challenge, which can implicitly bridge the distribution gap between the incomplete and the complete shapes. However, the global feature based adversarial learning framework may lose the detailed geometric information encoded in different levels of the network, which may lead to low completion quality. 
\begin{figure}[t]
    \centering
    \includegraphics[width=\linewidth]{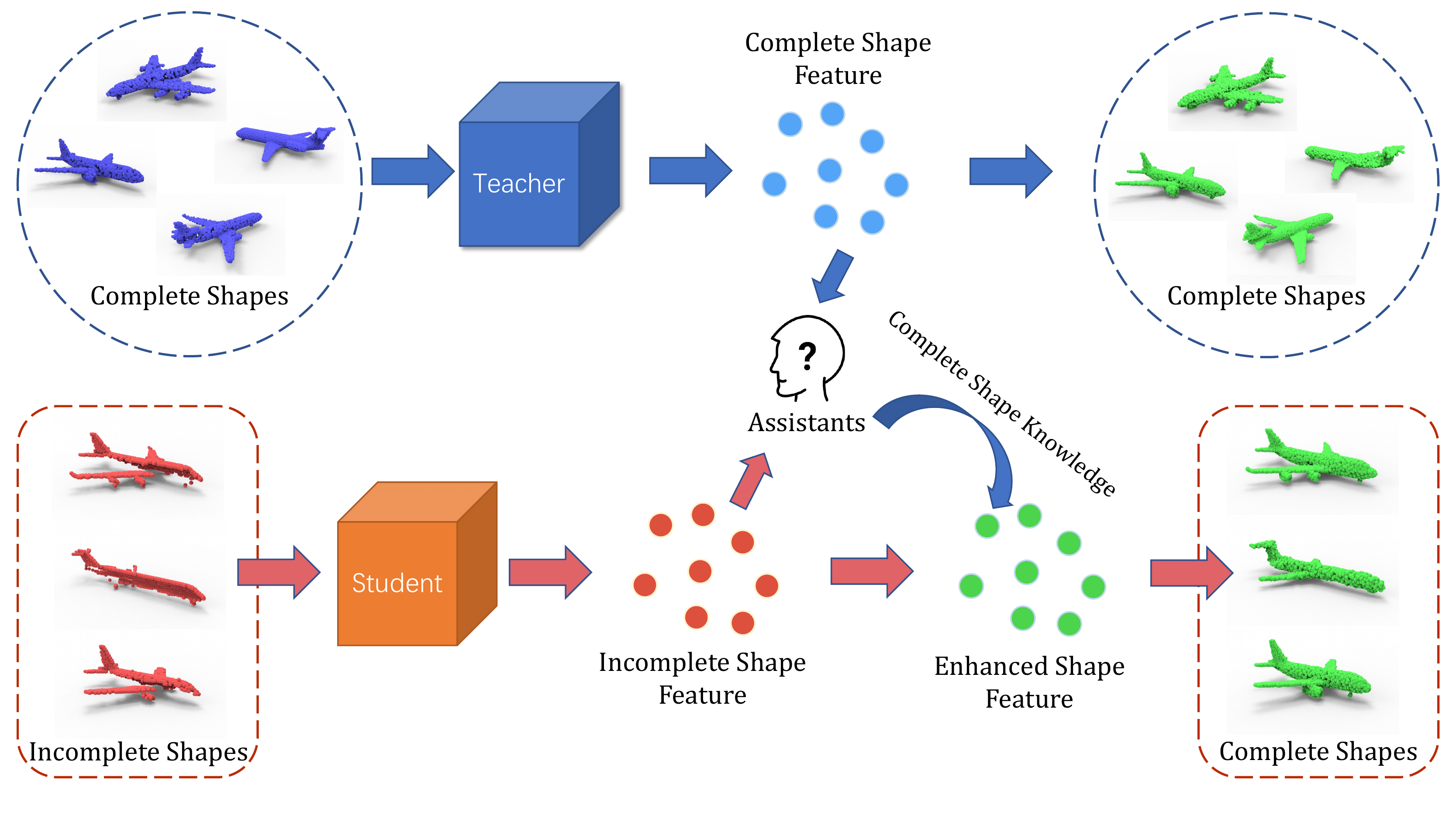}
    \caption{An illustration of our starting point. The teacher network is responsible for learning the knowledge of complete shape and transferring it to the student network through the assistant. The student network combines the specific incomplete shape and learns the knowledge of complete shape with the help of the assistants.}
    \label{fig: simplenet}
    \vspace{-0.6cm}
\end{figure}
On the other hand, implicitly aligning the feature can not fully explore the corresponding relationship between incomplete and complete shapes. Another representative work ShapeInversion \cite{zhang2021unsupervised} introduces GAN inversion for shape completion with the help of a well-pretrained generator and a proper latent code. But it takes much time to search for the appropriate latent code during the inference stage, and an inappropriate latent code might cause disaster completion results. 

Inspired by the great success of knowledge distillation \cite{hinton2015distilling}, we propose a new perspective to the unpaired shape completion task, which is how to transfer the knowledge from the complete shape domain into the incomplete shape domain. In this paper, the term "knowledge" specifically refers to the common representation in the latent space of the complete shape, which should be distilled to guide the completion process for incomplete input. This idea raises several questions: (1) How to implement a knowledge transfer process? (2) Where should the knowledge transfer take place? (3) How to evaluate the eﬀect of knowledge transfer? (4) How to restore high-fidelity shape after knowledge transfer? 

To solve the above problems, we design a novel Teacher-Assistant-Student framework as shown in Fig. \ref{fig: simplenet}, which is composed of a series of elaborating knowledge transfer processes. These processes are similar to schooling, where the teachers impart knowledge to assistants and students, and assistants instruct students to digest knowledge and evaluate the status of learning outcomes. 
A \emph{\textbf{Knowledge Transfer Implementation}}: The teacher network takes a complete shape as the input and learns the common knowledge of the complete shape by reconstructing it. The student network takes an incomplete shape as input and restores its complete shape with the guidance of the complete shape knowledge. To implement a knowledge transfer, the proposed Knowledge Recovery Assistant (KRA) inputs the incomplete shape features learned by the student network and adopts a residual module to explicitly infer and supplement the missing complete shape knowledge, thus enhancing the incomplete shape features by encoding the complete shape knowledge.
\emph{\textbf{Multi-Stage Knowledge Transfer}}: The knowledge in the different layers encodes irreplaceable geometric information of complete shape. Therefore, we propose a coarse-to-fine knowledge transfer strategy to sufficiently learn the multi-level knowledge of complete shape by iteratively employing the KRA module in different layers. 
\emph{\textbf{Effectiveness Evaluation of Knowledge Transfer}}: We propose a mini-GAN based Knowledge Discrimination Assistant (KDA) to evaluate the effectiveness of knowledge transfer. Specifically, the KDA accepts the complete shape feature from the teacher network as exemplary knowledge and discriminates against the enhanced features after knowledge transfer from the student network. 
\emph{\textbf{High-fidelity Complete Shape Restoration}}: We introduce a training strategy 
to restore the high-fidelity complete shapes. For the teacher network, the gradient shielding technology is adapted to ensure the gradient of incomplete shape does not ﬂow into the teacher network during backpropagation. For the student network, we share the restoration module structure and parameters of the teacher network, 
which makes it possible to restore high-fidelity complete shapes based on the enhanced features.

We conduct experiments on the common benchmarks of unpaired shape completion task, including synthetic datasets (3D-EPN dataset \cite{dai2017shape} and CRN dataset \cite{wang2020cascaded}) and real-world scans (MatterPort3D \cite{chang2017matterport3d}, ScanNet \cite{dai2017scannet} and KITTI \cite{geiger2012we}). The results demonstrate that our method outperforms the state-of-the-art (SOTA) unpaired shape completion methods.
Our main contributions are summarized as follows.
\begin{itemize}

\item We formulate the unpaired shape completion task as the knowledge transfer task, and design a novel end-to-end completion network composed of the paralleled teacher and student networks called \cznet~. Under the guidance of complete shape knowledge, the student network can restore a reasonable high-quality point cloud.
\item We introduce the Knowledge \Assistanta~and the Knowledge \Assistantb~to aid the student network in explicitly inferring and supplementing the missing complete shape knowledge, thus enhancing the incomplete shape feature by encoding the complete shape knowledge. 
\item We conduct comprehensive experiments on several datasets and show that our method can achieve SOTA performance over the previous unpaired shape completion methods.
\end{itemize}
\begin{figure*}[h]
\setlength{\abovecaptionskip}{-15pt}
\setlength{\belowcaptionskip}{-0pt}
  \centering
  \includegraphics[width=\linewidth]{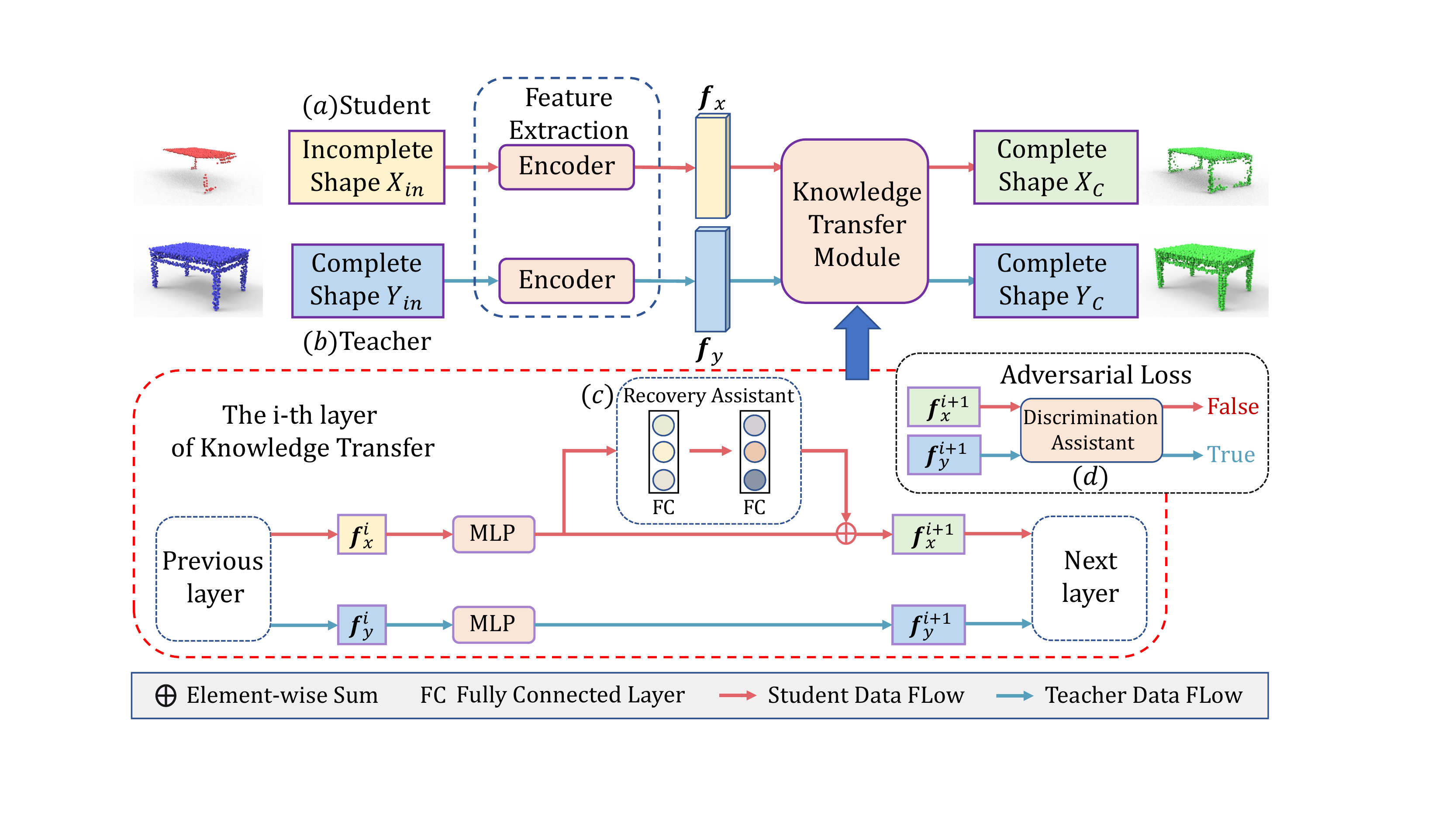}

  \caption{The overall architecture of \cznet, which consists of the paralleled teacher and student networks. (a) The above is the student network, which converts an incomplete shape into a complete one. (b) And the below is the teacher network, which reconstructs the complete shape and transfers the knowledge to the student network. (c) The Knowledge \Assistanta~(KRA) in the student network is adopted to supplement the missing complete shape knowledge to adapt to the task of point cloud completion. (d) The Knowledge \Assistantb~(KDA) is responsible for identifying which features come from the teacher network.}
  \label{fig:overview}
\end{figure*}
\section{Related Work}
\label{sec:formatting}
\noindent\textbf{Traditional Shape Completion Methods.} Traditional point cloud completion methods usually utilize the geometric attributes of objects \cite{hu2019local,mitra2006partial} or retrieval of the complete structure in the database \cite{han2008bottom,li2015database}. These methods usually leverage the artificially constructed features for completion, which cannot robustly generalize to the cases of complex 3D surfaces with large missing parts.

\noindent\textbf{3D Shape Completion with Pair Supervision.}
With the development of the deep neural network, researchers tend to leverage learning-based methods for 3D shape completion. Early methods are based on the 3D voxel grid \cite{yang20173d,wang2017shape}, but they are limited by the computational cost, which increases cubically to the shape resolution. 
On the other hand, point cloud based methods have emerged a lot in recent years, which is benefited from the pioneering work PointNet \cite{qi2017pointnet} of point cloud feature extraction and its subsequent studies \cite{qi2017pointnet++, wang2019dynamic}. 
PCN \cite{yuan2018pcn} is the first learning-based completion network that can directly operate on point clouds, which uses an encoder-decoder structure to predict complete point clouds.
More recent works \cite{han2019multi,zhang2020detail,wen2020point,richard2020kaplan, yu2021pointr, huang2021rfnet, gong2021me, xie2021style,wen2021pmp, xiang2021snowflakenet,alliegro2021denoise, zhang2021pc, xia2021asfm, Wang_2022_CVPR, mittal2022autosdf,  Yan_2022_CVPR} have made efforts to preserve the observed geometric details from the local features in incomplete inputs. 
VRC-Net \cite{pan2021variational} proposes a variational framework by leveraging the relationship between structures during the completion process.
PDR \cite{lyu2021conditional} introduces the conditional denoising diffusion probabilistic model to generate complete shapes.
LAKe-Net \cite{Tang_2022_CVPR} proposes a novel topology-aware point cloud completion model by localizing aligned keypoints, with a novel Keypoints-Skeleton-Shape prediction manner.
SeedFormer \cite{zhou2022seedformer} introduces a new shape representation, which not only captures general structures from partial inputs but also preserves regional information of local patterns.
There are also some networks using a voxel-based completion process. GRNet \cite{xie2020grnet} proposes a gridding network for dense point reconstruction. VE-PCN \cite{wang2021voxel} develops a voxel-based network for point cloud completion by leveraging edge generation.  

\noindent\textbf{3D Shape Completion without Pair Supervision.} On the other hand, it is difficult to obtain the paired ground truth of real-scans, so a few studies on unpaired shape completion has been proposed. 
AML \cite{stutz2018learning} uses the maximum likelihood method to measure the distance between complete and incomplete point clouds in the latent space. Pcl2Pcl \cite{chen2019unpaired} pretrains two auto-encoders, and directly learns the mapping from partial shapes to the complete ones in the latent space. Wu et al. \cite{wu2020multimodal} refer to the method of VAE to achieve multimodal outputs. MM-Flow \cite{zhao2021mm} proposes a flow-based network together with a multi-modal mapping strategy for 3D point cloud completion.  Cycle4Completion \cite{wen2021cycle4completion} designs two cycle transformations to establish the geometric correspondence between incomplete and complete shapes in both directions. ShapeInversion \cite{zhang2021unsupervised} incorporates a well-trained GAN as an effective prior for shape completion. Cai et al. \cite{cai2022learning} establish a unified and structured latent space to achieve partial-complete geometry consistency and shape completion accuracy.
\section{Method}
\label{sec:formatting}
\subsection{Overview}
Our \cznet~intends to transfer the knowledge from complete shape to incomplete shape based on two paralleled teacher and student networks. Fig. \ref{fig:overview} shows the overall framework of our method. $X_{in}$ and $Y_{in}$ denote the incomplete and the complete point clouds, respectively. Note that for unpaired completion, there is no correspondence between $Y_{in}$ and $X_{in}$. The teacher network takes $Y_{in}$ as input and learns the knowledge of complete shape by reconstructing the same complete shape $Y_c$ as $Y_{in}$. The student network takes $X_{in}$ as input and expects to restore the complete prediction $X_c$. Feature extraction module is responsible for extracting the global features $\emph{\textbf{f}}_x$ and $\emph{\textbf{f}}_y$ from the input point cloud $X_{in}$ and $Y_{in}$. The knowledge transfer module accepts the features extracted from both networks, and then guides the student network to explicitly infer and supplement the missing complete shape knowledge with the help of our proposed assistants, thus enhancing the incomplete shape features by encoding the complete shape knowledge. Finally, a unified MLP is used to convert the enhanced features into complete shapes.
\subsection{Feature Extraction Module}
With a given point set $P \in R^{N \times 3},$ our encoder extracts a k-dimensional global feature $\emph{\textbf{f}} \in R^k$.
Similar to PCN, the encoder consists of MLP layers for point-wise feature extraction and finally performs maxpooling operation to obtain the global feature.

\subsection{Knowledge Transfer Module}

This module takes the k-dimensional latent features $\emph{\textbf{f}}_x$ and $\emph{\textbf{f}}_y$ from the student and teacher networks as input, which can promote the student network to learn the complete shape knowledge from the teacher network. 

\noindent\textbf{KRA for the Knowledge Transfer.} For the teacher network, 
the feature $\emph{\textbf{f}}_y^{i}$ of the $i$-th layer contains the knowledge of complete shape, which are gradually pushed from high dimensional latent space to Euclidean space.
For the student network, the KRA is introduced to enhance the incomplete shape features by explicitly learning and encoding the complete shape knowledge.
In our design, the knowledge of the missing region is inferred and supplemented through residual connection. As shown in Fig. \ref{fig:overview},
we select a simple but effective model to accomplish this work. 
\begin{equation}
\label{eq:g}
    KRA(x)= FC\circ ReLU\circ FC(x),
\end{equation}
where $FC$ means a fully connected layer and $ReLU$ is an activation function.

The knowledge in each stage is irreplaceable, and the shallow knowledge is assumed to be more specific than the deep one. To enhance 3D shape understanding ability, we propose a multi-stage transfer strategy by iteratively employing the KRA module in different layers, which contributes to the mastery of more comprehensive complete shape knowledge for the student network via a continuous learning procedure. 

\noindent\textbf{KDA for the Effect Evaluation of Knowledge Transfer.} 
An additional assistant KDA that is applied in each layer is adopted to receive the complete shape knowledge from the teacher network, and evaluate the learning effectiveness of the student on this basis. However, in the unpaired task, it is unreasonable to directly transfer the unprocessed complete shape knowledge to the incomplete shape of the same batch due to the missing one-to-one correspondence. Considering this, and in order to avoid over complex knowledge transfer design, the KDA employs the generative adversarial architecture to make the student network automatically obtain the knowledge from the teacher network. Specifically, the KDA regards the feature from the teacher network as the representative of the complete shape knowledge and assigns it to $True$, while the enhanced feature from the student network is assigned to $False$. Under the guidance of the KDA, the student network strives to make its enhanced feature mimic the one from the teacher network. The complete shape knowledge has been transferred from the teacher network to the student network when the KDA can hardly distinguish the two types of features.

The discrimination loss can be defined as follows:
\begin{equation}
\label{eq:LD}
    \mathcal{L}_D=\frac{1}{2n}\sum_{i=1}^n\left\{{E}_x [D_i(\emph{\textbf{f}}_x^i)]^2 + {E}_y [D_i (\emph{\textbf{f}}_y^i)-1]^2 \right\}.
\end{equation}
where $D_i$ represents the $i$-th KDA, ${E}$ denotes the mathematical expectation, $\emph{\textbf{f}}_x^i$ represents the feature of the student network at the $i$-th layer of the knowledge transfer module, $\emph{\textbf{f}}_y^i$ is the feature of the teacher network at the $i$-th layer and $n$ is the number of layers of the knowledge transfer module. 
On the other hand, the generated loss used to evaluate the student network's learning achievements is defined as:
\begin{equation}
\label{eq:match}
    \mathcal{L}_{G}=\frac{1}{n}\sum_{i=1}^n{E}_x [D_i(\emph{\textbf{f}}_x^i)-1]^2.
\end{equation}

\subsection{Point Cloud Restoration} To restore the complete shapes from the features, an MLP is applied to map the feature $\emph{\textbf{f}}$ from $R^k$ to $R^{3N}$, where $N$ is the number of generated points. Then the feature is reshaped from $R^{3N}$ to $R^{N \times 3}$ to get the point cloud $P$. 

The teacher network follows an auto-encoder architecture for self-reconstruction. After obtaining the reconstructed complete shape $Y_c$, 
we leverage a common permutation-invariant metrics Earth Mover’s Distance (EMD) to supervise the training, which is shown below.
\begin{equation}
\label{eq:recon}
    \mathcal{L}_{teacher}=\mathcal{L}_{EMD}(Y_{in},Y_{c}).
\end{equation}

For the student network, the enhanced feature is mapped to the complete shape $X_c$ by the shared MLP in the teacher network.
Nevertheless, there is no corresponding ground truth for the predicted complete shape $X_c$, which indicates a direct implementation of the bi-direction EMD may not yield desirable results. Therefore, in order to prevent the mode collapse and preserve the input information, we apply Unidirectional Chamfer Distance($UCD$) 
and the loss of the student network is expressed as:
\begin{equation}
\label{eq:com}
    \mathcal{L}_{student}=\mathcal{L}_{UCD}(X_{in},X_{c}).
\end{equation}
\begin{table*}[h]
\setlength{\abovecaptionskip}{-0pt}
\setlength{\belowcaptionskip}{-0pt}
\caption{Shape completion results on 3D-EPN dataset. The numbers shown are $CD$ (lower is better), which is scaled by $10^4$. \cznet~outperforms other unpaired methods by a large margin, and is comparable to the various paired methods.}
\label{tab:3depn}
\begin{center}
    \begin{tabular}{c|c|c|cccccccc}
    \toprule
  & Methods & Average & Plane & Cabinet & Car & Chair & Lamp & Sofa & Table & Boat \\
    \midrule
  $paired$ & 3D-EPN \cite{dai2017shape} & 29.1 & 60.0 & 27.0 & 24.0 & 16.0 & 38.0 & 45.0 & 14.0 & 9.0 \\
  & FoldingNet \cite{yang2018foldingnet} & 9.2 & 2.4 & 8.5 & 7.2 & 10.3 & 14.1 & 9.1 & 13.6 & 8.8 \\
  & PCN \cite{yuan2018pcn} & \textbf{7.6} & \textbf{2.0} & \textbf{8.0} & \textbf{5.0} & 9.0 & 13.0 & 8.0 & 10.0 & \textbf{6.0} \\
  & TopNet \cite{tchapmi2019topnet} & 8.4 & 2.5 & 8.8 & 5.9 & 9.3 & 12.0 & 8.4 & 13.5 & 7.1 \\
  & SA-Net \cite{wen2020point} & 7.7 & 2.2 & 9.1 & 5.6 & \textbf{8.9} & \textbf{10.0} & \textbf{7.8} & \textbf{9.9} & 7.2 \\
    \hline
  $unpaired$ & AE(baseline) \cite{chen2019unpaired} & 25.4 & 4.0 & 37.0 & 19.0 &31.0 & 26.0 &30.0 & 44.0 & 12.0 \\
  & Pcl2Pcl \cite{chen2019unpaired} &17.4 & 4.0 & 19.0 & 10.0 & 20.0 & 23.0 & 26.0 & 26.0 & 11.0 \\
  & Cycle4Completion \cite{wen2021cycle4completion} & 14.3 & 3.7 & 12.6 & 8.1 & 14.6 & 18.2 & 26.2 & 22.5 & 8.7 \\
  & Ours &\textbf{10.0} & \textbf{2.6} & \textbf{10.9}& \textbf{6.3} &\textbf{12.4} & \textbf{15.1} & \textbf{10.5} &\textbf{15.8} &\textbf{6.5} \\
  \bottomrule
  \end{tabular}
\end{center}
\end{table*}
\begin{table*}[h]
\setlength{\abovecaptionskip}{-0pt}
\setlength{\belowcaptionskip}{-0pt}
\caption{Shape completion results on CRN dataset. The numbers shown are $CD$ (lower is better), which is scaled by $10^4$. \cznet~outperforms other unpaired methods by a large margin, and is comparable to the various paired methods.}
\label{tab: CRN}
\begin{center}
    \begin{tabular}{c|c|c|cccccccc}
    \toprule
  & Methods & Average & Plane & Cabinet & Car & Chair & Lamp & Sofa & Table & Boat \\
    \midrule
  $paired$ & PCN \cite{yuan2018pcn} & 9.1 & 3.5 & \textbf{11.3} & 6.4 & 11.0 & 11.6 & 11.5 & 10.4 & 7.4 \\
  & TopNet \cite{tchapmi2019topnet} & 11.4 & 4.1 & 12.9 & 7.8 & 13.4 & 14.8 & 16.0 & 12.9 & 8.9 \\
  & MSN \cite{liu2020morphing} & 8.8 & 2.9 & 12.5 & 7.1 & 10.6 & 9.3 & 12.0 & 9.6 & 6.5  \\
  & CRN \cite{wang2020cascaded} & \textbf{8.0}  & \textbf{2.3} & 11.4 & \textbf{6.2} & \textbf{8.8} & \textbf{8.5} & \textbf{11.3} & \textbf{9.3} & \textbf{6.1} \\
    \hline
  $unpaired$ & Pcl2Pcl \cite{chen2019unpaired}  & 22.4 & 9.8 & 27.1 & 15.8 & 26.9 & 25.7 & 34.1 & 23.6 & 15.7 \\
  & ShapeInversion \cite{zhang2021unsupervised}& 14.9 & 5.6 & 16.1 & 13.0 & 15.4 & 18.0 & 24.6 & 16.2 &10.1  \\
  & Ours &\textbf{11.5} &\textbf{3.8}& \textbf{12.4}&\textbf{9.0} &\textbf{13.9}  & \textbf{14.1} & \textbf{18.1} &\textbf{11.7} & \textbf{8.7} \\
  \bottomrule
    \end{tabular}
\end{center}
\end{table*}
\subsection{Training Strategy}
\label{sec: training}
Since the teacher and student networks share the restoration module, the quality of the student network's completion results depends on the teacher network's ability to reconstruct complete shapes.
To enable the teacher network to reconstruct a high-quality complete shape, we adopt a special training strategy, which ensures the gradient of incomplete shape does not flow into the teacher network at the knowledge transfer module during backpropagation.

We use $FE$ and $KRA$ to represent the feature extraction module and the knowledge \assistanta s in the knowledge transfer module, respectively. In particular, $KT$ denotes the remaining structure of the knowledge transfer module. 
In our model, there are three types of loss functions. $\theta_D$ is used to represent the trainable parameters in $\left\{D_i |i=1,...,k\right\}$ (i.e. all the knowledge discrimination assistants). $\theta_{teacher}$ is adopted to stand for the trainable parameters in $\left\{FE, KT\right\}$. And $\theta_{student}$ denotes the trainable parameters in $\left\{FE, KRA\right\}$.

For a given learning rate $\gamma$, we first update $\theta_D$ of all the discriminators and freeze all other parts, where the gradient optimization step is expressed as:
\begin{equation}
\label{eq:thetaD}
    \theta_D \leftarrow \theta_D - 2\gamma \frac{\partial\mathcal{L}_D}{\partial\theta_D}.
\end{equation}
Then we use $\mathcal{L}_{teacher}$ to regularize the teacher network. The gradient descent step is shown below.
\begin{equation}
\label{eq:thetaRecon}
    \theta_{teacher}\leftarrow\theta_{teacher}-\gamma\frac{\partial\mathcal{L}_{teacher}}{\partial\theta_{teacher}}.
\end{equation}
To prevent the teacher network from generating bad reconstructed complete shapes, $\mathcal{L}_{student}$ and $\mathcal{L}_{G}$ are only adopted to update $\theta_{student}$. Such a strategy enables the teacher network only to learn the complete shape knowledge, thus reconstructing a high-quality point cloud.
\begin{equation}
\label{eq:thetaCom}
    \theta_{student} \leftarrow \theta_{student} - \gamma [\lambda_g\frac{\partial\mathcal{L}_{G}}{\partial\theta_{student}} + \lambda_p\frac{\partial\mathcal{L}_{student}}{\partial\theta_{student}}],
\end{equation}
where $\lambda_g$ and $\lambda_p$ are hyper-parameters and we set $\lambda_g=0.1$ and $\lambda_p=1$.

\noindent\textbf{Inference Strategy.} The teacher network and the KDA transfer the complete shape knowledge to the student network and are used in the training stage only. During the inference, like the student data flow as shown in Fig. \ref{fig:overview}, only the student network and the KRA are adopted, and the incomplete shape is fed to the student network to get the predicted complete shape.
\section{Experiments}
\noindent\textbf{Datasets.} We evaluate our network on two systhetic datasets 3D-EPN \cite{dai2017shape} and CRN \cite{wang2020cascaded}, and three real-word datasets MatterPort3D \cite{chang2017matterport3d}, ScanNet \cite{dai2017scannet} and KITTI \cite{geiger2012we}.

\noindent\textbf{Evaluation Metrics.}  For evaluation on synthetic datasets, we use Chamfer Distance($CD$) to compare the difference between the predicted point cloud and the ground truth.
$CD$ can be defined as:
\begin{equation}
\label{eq:cdt}
    \mathcal{L}_{CD}(P,Q)=\frac{1}{N_P}\sum_{p\in P}\min_{q \in Q} ||p-q||_2^2+
    \frac{1}{N_Q}\sum_{q\in Q}\min_{p \in P}||p-q||_2^2.
\end{equation}
For the real-world datasets without ground truth, there is not a very appropriate way to evaluate the prediction point cloud. We use Unidirectional Hausdorff Distance($UHD$) to evaluate the similarity between the input point cloud and the predicted point cloud.
$UHD$ can be defined as:
\begin{equation}
\label{eq:ucdt}
    \mathcal{L}_{UHD}(P,Q)=\max_{p\in P}\min_{q \in Q} ||p-q||_2^2.
\end{equation}

\noindent\textbf{Implementation Details.} We follow the previous unpaired shape completion methods \cite{chen2019unpaired, wen2021cycle4completion, zhang2021unsupervised} to train a separate network for each category. The number of points of the complete shapes is 2048 for all datasets. We set the learning rate to $10^{-4}$, the batch size to 32 and the epochs to 600 during the training.
\subsection{Evaluation on 3D-EPN Dataset}

To make a fair comparison with the previous unpaired shape completion methods (i.e., Pcl2Pcl \cite{chen2019unpaired} and Cycle4Completion \cite{wen2021cycle4completion}), we use 3D-EPN dataset \cite{dai2017shape} with the same training and testing split for evaluation. Note that there is no correspondence between incomplete and complete shapes during training.  
The quantitative results are shown in Tab. \ref{tab:3depn}, where we use Chamfer Distance(CD) as the evaluation metric.
The experimental results show that our \cznet~ achieves the best results over the counterpart unpaired completion methods. Especially, compared with the second-best method Cycle4Completion, \cznet~ reduces the CD loss with a non-trivial margin by 30\%, and achieves the best results in terms of CD loss over all categories. In our opinion, the reason that \cznet~ achieves better performance over Cycle4Completion can be dedicated to the more detailed geometric recovery ability by the proposed two assistant modules. Moreover, when compared with the supervised methods, \cznet~ also yields comparable results, which is quite close to the CD loss of FoldingNet \cite{yang2018foldingnet}. In Fig. \ref{fig: 3depn}, we visually compare the performance of \cznet~ with the other unpaired completion methods, from which we can find that \cznet~ can predict more accurate and detailed shapes in the missing regions, and generate a better completion quality. For example, in the fifth column of Fig. \ref{fig: 3depn}, \cznet~ successfully predicts the correct shape of table legs, while the other two methods (i.e. Pcl2Pcl and Cycle4Completion) failed to reconstruct the same legs as ground truth.
\subsection{Evaluation on CRN Dataset} To verify the effectiveness of our network on different datasets and make a fair comparison with ShapeInversion \cite{zhang2021unsupervised}, we also conducted an evaluation on CRN datasets \cite{wang2020cascaded}. The quantitative results are shown in Tab. \ref{tab: CRN}. It can be seen that our model outperforms the second-best unpaired method ShapeInversion with a large margin by more than 20\%, and achieves the best results in all the classes, which proves the superiority of our method. Besides the disadvantage of a long inference time, ShapeInversion pays excessive attention to the input information and ignores the overall shape, which limits its performance. And our \cznet~ can reasonably encode the complete shape knowledge into the incomplete shape feature, thus generating a high-quality point cloud. 
The visual comparison with other unpaired shape completion methods is shown in Fig. \ref{fig: crn} and we can observe that \cznet~ can predict more detailed and smooth complete shapes. For instance, in the third column in Fig. \ref{fig: crn}, the predicted chair of ShapeInversion does not represent the same object as the ground truth, while the one of \cznet~ has a reasonable overall shape.
\begin{table}[H]
\setlength{\abovecaptionskip}{-0pt}
\setlength{\belowcaptionskip}{-0pt}
\caption{Shape completion results on the real scans. The numbers shown are [$UHD$](lower is better), where $UHD$ is scaled by $10^2$.}
\label{tab: realscan}
\begin{center}
\resizebox{\linewidth}{!}{ 
    \begin{tabular}{c|c|c|c|c|c}
    \hline
   \multirow{2}{*}{Methods}  & \multicolumn{2}{c}{ScanNet} & \multicolumn{2}{c}{MatterPort3D} & KITTI  \\
   \cline{2-6}
 & Chair & Table &Chair & Table & Car\\
 \cline{1-6}
Pcl2Pcl  & 10.1 & 11.8 & 10.5 & 11.8 & 14.1 \\
\hline
ShapeInversion  & 10.1 & 11.9 & 10.0 &11.8 & 13.8  \\
\hline
Ours & \textbf{6.4} &  \textbf{6.3} & \textbf{8.4} & \textbf{10.0} & \textbf{6.0} \\
\hline
    \end{tabular}}
\end{center}
\end{table}
\subsection{Evaluation on Real-World Scans}
\begin{figure}[H]
    \centering
    \includegraphics[width=\linewidth]{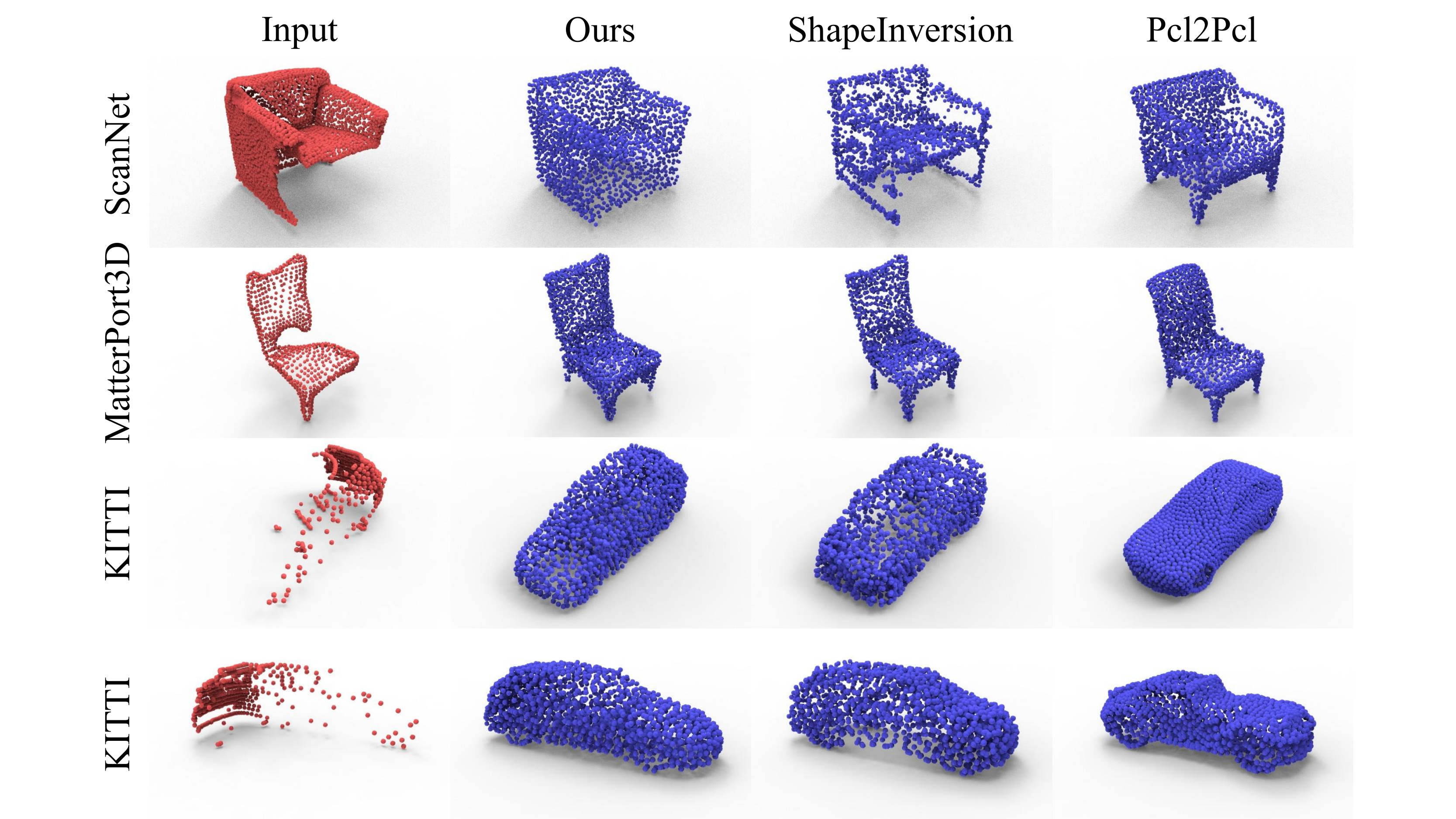}
    \caption{Shape completion on real-world partial scans. }
    \label{fig: realscan}
\end{figure}

We further evaluate the generalization ability of our network on real-world data extracted from MatterPort3D, ScanNet and KITTI. Our network trained on CRN dataset can be directly generalized to the real-world scans from the datasets mentioned above without further fine-tuning process. 
Quantitative and qualitative results shown in Tab. \ref{tab: realscan} and Fig. \ref{fig: realscan}, respectively. Even in the case of severe sparseness and incompleteness (such as cars in KITTI), our network can predict a reasonable complete point cloud. Quantitative results also show our method can well preserve the input information rather than predict an average shape of this category only.
For example, in the first line of Fig. \ref{fig: realscan}, \cznet~predicts reasonable armrest and legs of chair, while the results of other methods may not represent the same object as the input.

\begin{figure*}
\setlength{\abovecaptionskip}{-0pt}
\setlength{\belowcaptionskip}{-0pt}
\centering
\begin{subfigure}{0.85\linewidth}{
\setlength{\abovecaptionskip}{-0pt}
\setlength{\belowcaptionskip}{-0pt}
\includegraphics[width=\linewidth]{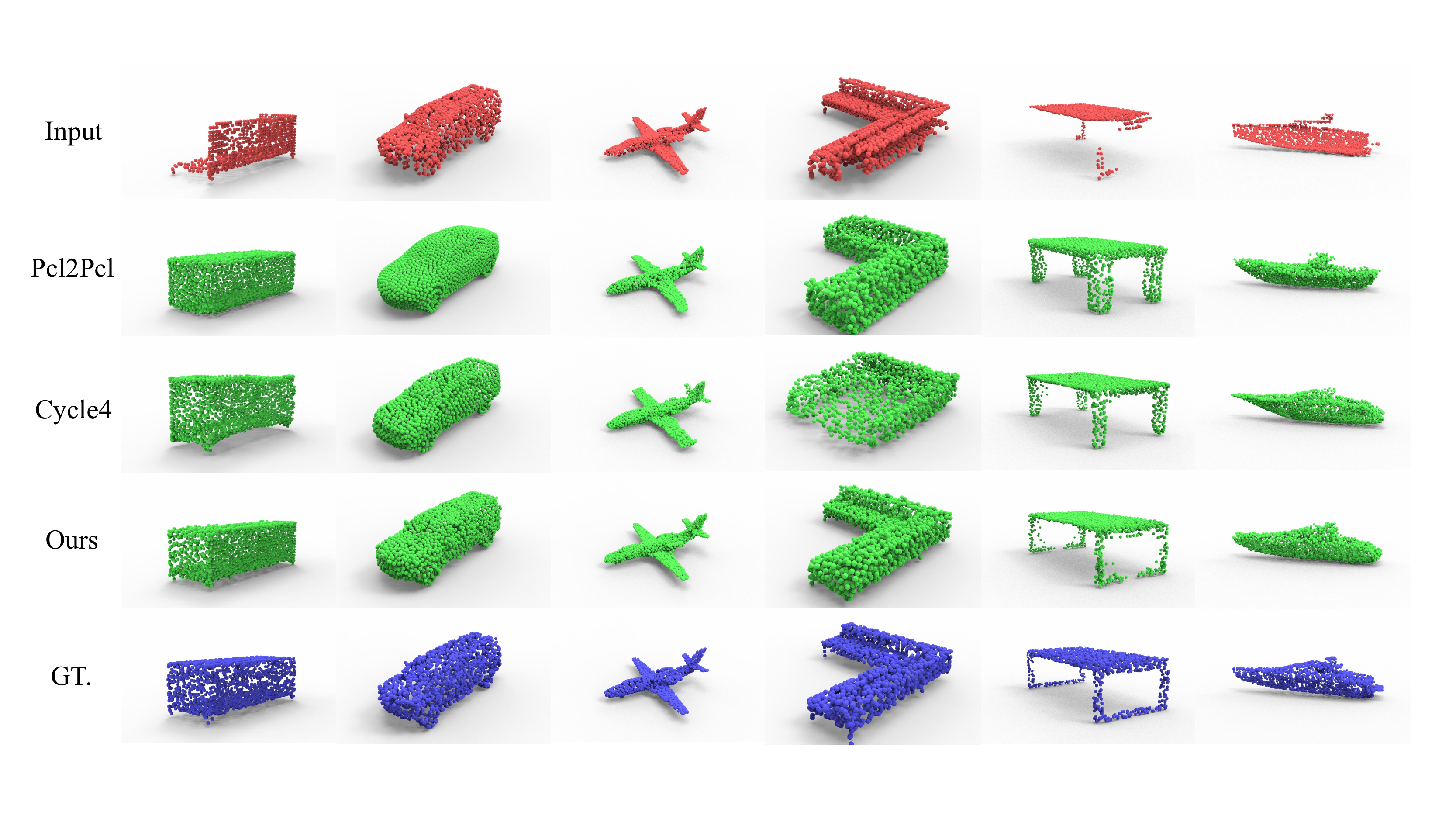}
\caption{Visual comparison with the state-of-the-art unpaired shape completion methods on 3D-EPN dataset.}
\label{fig: 3depn}
}
\end{subfigure}

\begin{subfigure}{0.85\linewidth}{
\setlength{\abovecaptionskip}{-0pt}
\setlength{\belowcaptionskip}{-0pt}
\includegraphics[width=\linewidth]{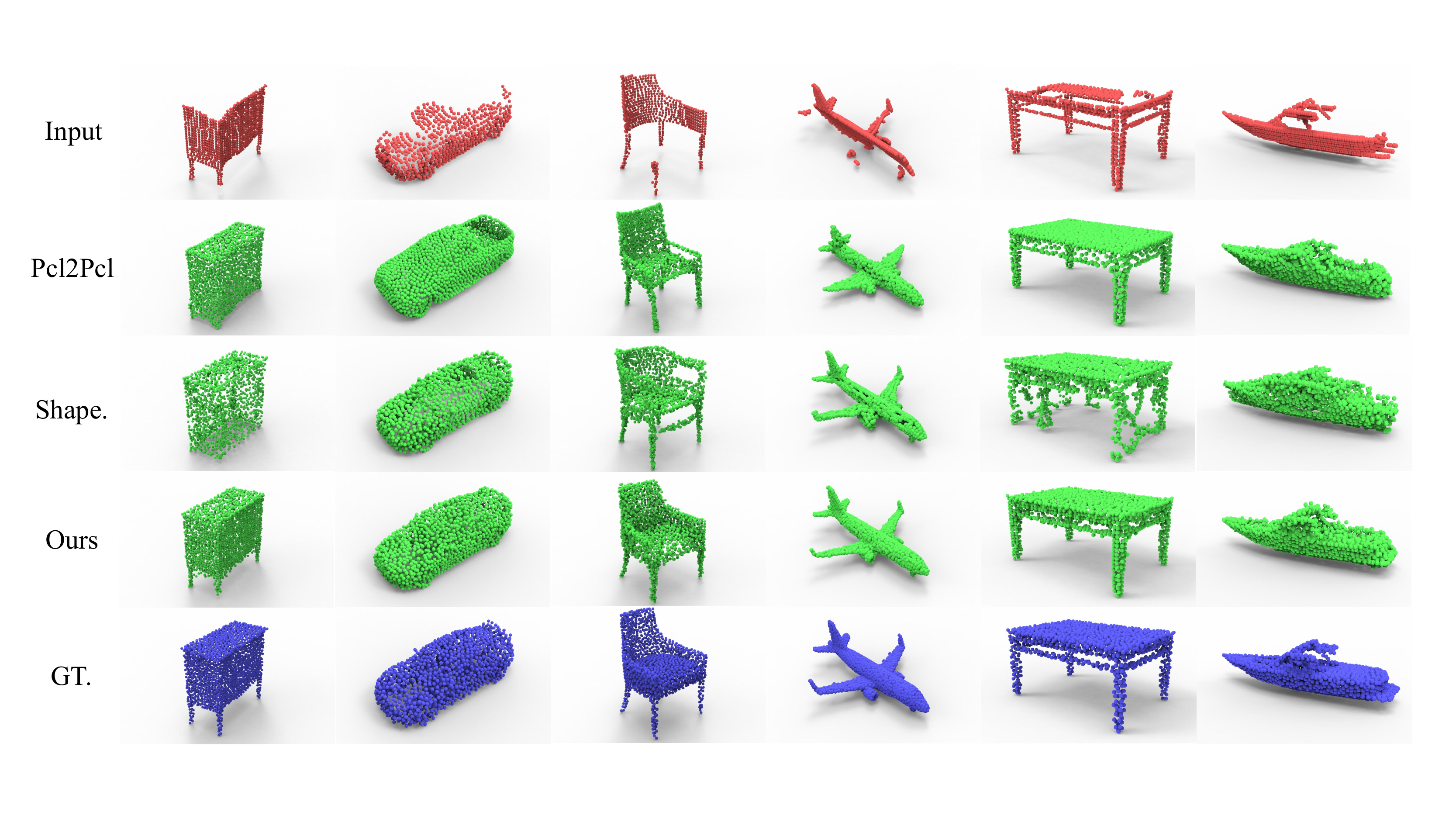}
\caption{Visual comparison with the state-of-the-art unpaired shape completion methods on CRN dataset.}
\label{fig: crn}
}
\end{subfigure}
\caption{Evaluation on synthetic datasets.}
\end{figure*}
\subsection{Analysis of Different Training Strategies}
In our model, the student loss $\mathcal{L}_{student}$ and the adversarial loss $\mathcal{L}_{G}$ are used to update $\theta_{student}$ only. However, $\mathcal{L}_{student}$ and $\mathcal{L}_{G}$ involve not only the knowledge \assistanta s but also the remaining part of the knowledge transfer module. To evaluate the effectiveness of other potential training strategies, similar to Sec. \ref{sec: training}, we use $\theta_{all}$ to denote the trainable parameters in \{$FE,KT,KRA$\}. Then we develop the variations of (a) $\partial\mathcal{L}_{student}/\partial\theta_{all}$ and (b) $\partial\mathcal{L}_{G}/\partial\theta_{all}$ to replace $\partial\mathcal{L}_{student}/\partial\theta_{student}$ or $\partial\mathcal{L}_{G}/\partial\theta_{student}$ in Eq. \ref{eq:thetaCom}.

The results are reported in Tab. \ref{tab: training}. We observed dropping performances in the variations of (a) and (b), where the teacher network is affected by the incomplete shape knowledge, which leads to a failure to reconstruct the complete shape input $Y_{in}$.
\begin{table}[h]
\setlength{\abovecaptionskip}{-0pt}
\setlength{\belowcaptionskip}{-0pt}
\caption{Results of different training strategies ($CD\times 10^4$).}
\label{tab: training}
\begin{center}
\resizebox{\linewidth}{!}{ 
    \begin{tabular}{l|c|ccccc}
    \toprule
   Methods & Average & Car & Chair & Lamp & Sofa & Table \\
   \midrule
$\partial\mathcal{L}_{com}/\partial\theta_{all}$  &22.9& 9.8 &22.3&27.3&30.1&25.0\\
$\partial\mathcal{L}_{match}/\partial\theta_{all}$ & 21.7& 10.5 & 22.0 & 26.1&32.3 &17.5 \\
\midrule
Original Model  & \textbf{13.4}& \textbf{9.0} & \textbf{13.9} & \textbf{14.1}&\textbf{18.1} &\textbf{11.7}  \\
  \bottomrule
    \end{tabular}}
\end{center}
\end{table}

\begin{figure*}
\setlength{\abovecaptionskip}{5pt}
\setlength{\belowcaptionskip}{-0pt}
\centering
\begin{subfigure}{0.49\linewidth}{
\setlength{\abovecaptionskip}{-5pt}
\setlength{\belowcaptionskip}{-5pt}
\includegraphics[width=\linewidth]{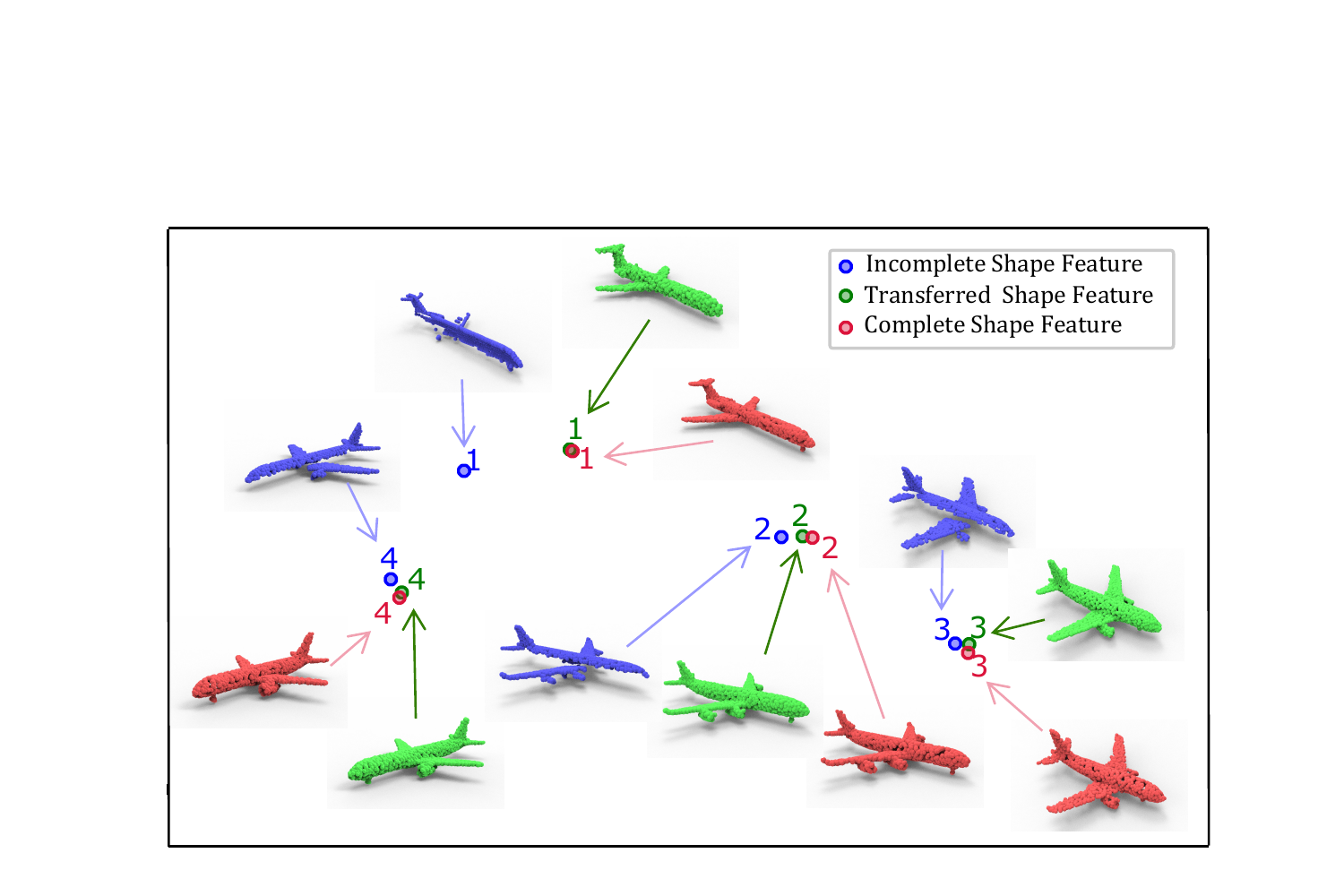} 
\caption{Visualization of latent space in our model.}
\label{fig:ours}}
\end{subfigure}
\begin{subfigure}{0.49\linewidth}{
\setlength{\abovecaptionskip}{-5pt}
\setlength{\belowcaptionskip}{-5pt}
\includegraphics[width=\linewidth]{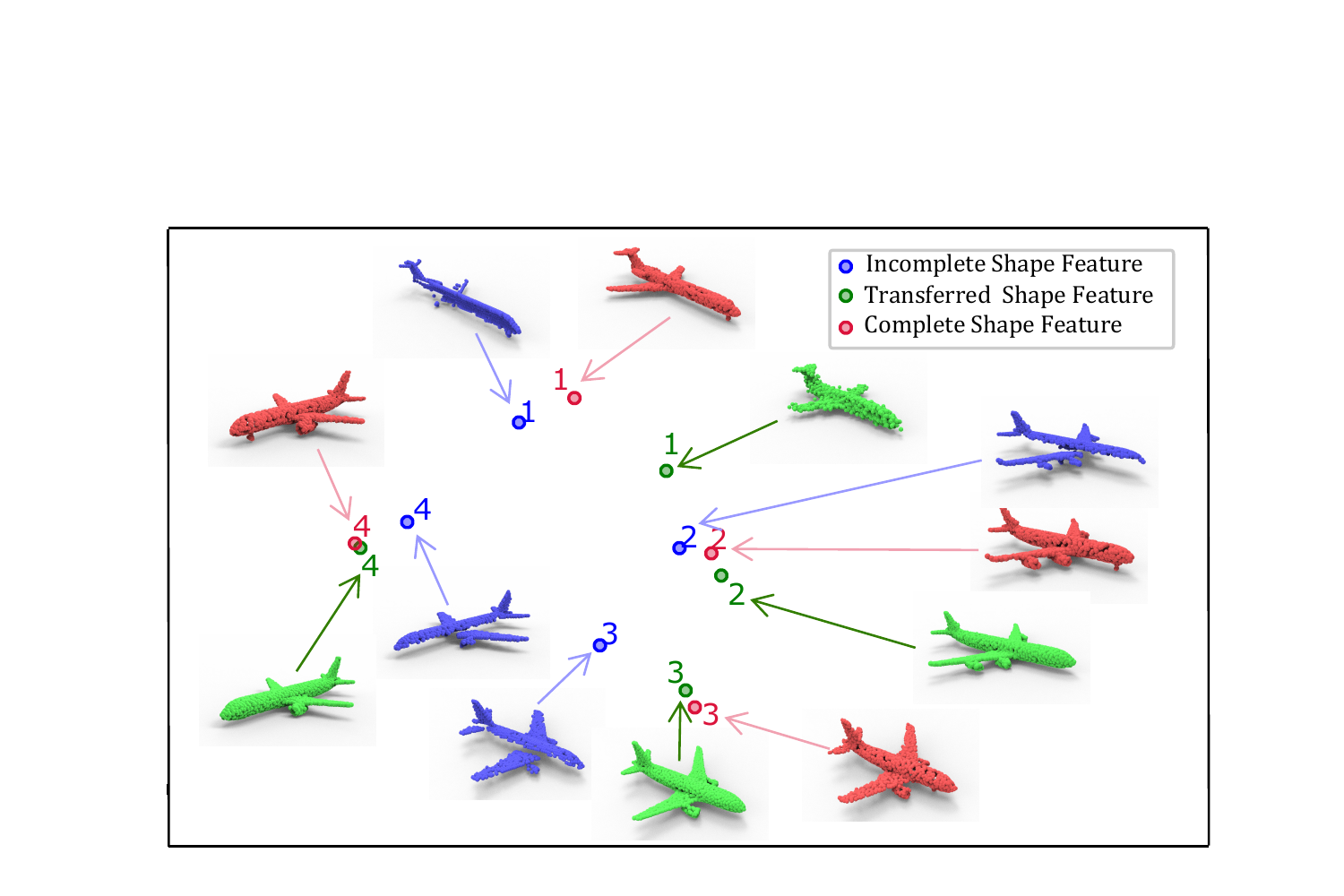} 
\caption{Visualization of latent space in Pcl2Pcl.}
\label{fig:pcl2pcl}}
\DeclareGraphicsExtensions.
\end{subfigure}
\caption{Visualization of latent space. (a) shows the results of our model, (b) shows the results of Pcl2Pcl. We further visualize the shape that these points represent.}
\label{fig: latent}
\end{figure*}

\subsection{Visual Analysis of Latent Feature Distribution}
In Fig. \ref{fig: latent}, we use t-SNE \cite{van2008visualizing} to visualize the features before and after the knowledge transfer module in the high-dimensional latent space as 2D points. The distance of points can reflect the similarity of the corresponding latent features to a certain extent. The red dots represent the features from complete shapes and the blue dots represent the features from the incomplete shapes before 
the knowledge transfer module. The green dots represent the transferred features of the incomplete shape after the knowledge transfer module. 

As shown in Fig. \ref{fig:ours}, the features from the teacher and student networks before the knowledge transfer module are not exactly the same, indicating that the correspondences in the incomplete shape and the complete shape have not been fully explored. After our knowledge transfer module, the features could be similar enough to adapt to the task of point cloud completion. We also visualize the features of Pcl2Pcl \cite{chen2019unpaired} on the same objects, as shown in Fig. \ref{fig:pcl2pcl}. Pcl2Pcl tends to output a reasonable complete shape without paying attention to maintaining the information of the incomplete shape input, resulting in the fact that its transformed features are still far away from those of the corresponding ground truth. Furthermore, we visualize the shape represented by these points, including the incomplete inputs, ground truths and predicted complete shapes.
\subsection{More Ablation Studies}
In this chapter, we further analyze each part of our network with results shown in Tab. \ref{tab: Analysis}. We perform this evaluation on five categories in CRN dataset \cite{wang2020cascaded}.

\noindent\textbf{\Assistanta~ Evaluation.} To evaluate the effectiveness of the components in our framework, we give a study about removing the KRA. In this case, we retrained the network for evaluation and the results are shown as \emph{w/o KRA} in Tab. \ref{tab: Analysis}. The network performance decreases without KRA, showing its effectiveness. 

\noindent\textbf{\Assistantb~ Evaluation.} The KDA is used to transfer the complete shape knowledge from the teacher network to the student network and judge the learning effectiveness of the student network. In order to evaluate the necessity of introducing the KDA, we retrained the network without using it and the results are shown as \emph{w/o KDA} in Tab. \ref{tab: Analysis}. The student network without the guidance of the KDA cannot learn the knowledge of complete shape, which leads to poor completion results.

\noindent\textbf{Residual Connection Evaluation.} In \cznet, we use the residual connection to reduce the learning pressure of the student network. We try to test the performance of the network without residual connection. The result is shown as \emph{w/o Residual}, which shows that residual connection can significantly improve the ability of the network. 

\noindent\textbf{Evaluation of $\mathcal{L}_{student}$.} $\mathcal{L}_{student}$ (shown in Eq. \ref{eq:com}) uses incomplete shape input $X_{in}$ to constrain the the complete shape output $X_{c}$, which makes $X_{in}$ and $X_{c}$ represent the same object. We try to train the network without $\mathcal{L}_{student}$ and the result is shown as \emph{w/o $\mathcal{L}_{student}$}.

\noindent\textbf{Comparison with transforming the global feature only.}
In order to explore the performance of our model that only transforms the global features before the knowledge transfer module between the two networks, we evaluate \cznet~under this circumstance. The result is shown as \emph{Only Global Feature} in Tab. \ref{tab: Analysis}.

\begin{table}[h]
\setlength{\abovecaptionskip}{-0pt}
\setlength{\belowcaptionskip}{-0pt}
\caption{More Analysis ($CD\times 10^4$).}
\label{tab: Analysis}
\begin{center}
\resizebox{\linewidth}{!}{ 
    \begin{tabular}{l|c|ccccc}
    \toprule
   Methods & Average & Car & Chair & Lamp & Sofa & Table \\
   \midrule
w/o KRA  &19.7& 10.9 &20.4&19.1&30.5&17.6\\
w/o KDA  &22.1& 10.7 &17.3&18.0&51.0&13.3\\
w/o Residual & 16.2 & 9.4 & 17.6 & 17.4 & 20.2 & 16.5 \\
w/o $\mathcal{L}_{student}$  &22.9& 9.8 &22.3&27.3&30.1&25.0\\
Only Global Feature & 19.6& 10.4 & 18.0 & 17.3&39.8 &12.5 \\
\midrule
Original Model  & \textbf{13.4}& \textbf{9.0} & \textbf{13.9} & \textbf{14.1}&\textbf{18.1} &\textbf{11.7}  \\
  \bottomrule
    \end{tabular}}
\end{center}
\end{table}
\section{Conclusions}
We propose the \cznet~for unpaired point cloud completion task. Compared with the previous methods, our method pays more attention to the process of knowledge transfer, and uses a teacher-student network combined with ingenious design to enhance the student network's understanding of shape. Furthermore, our model is evaluated on several widely used datasets, achieving state-of-the-art performance by a large margin compared with other unpaired completion methods. As our approach still needs unpaired supervision, we plan to explore the completion task in a fully unsupervised manner as future work.
\newpage
{\small
\bibliographystyle{ieee_fullname}
\bibliography{kt}
}

\end{document}